\documentclass[conference]{IEEEtran}
\IEEEoverridecommandlockouts
\usepackage{cite}
\usepackage{amsmath,amssymb,amsfonts}
\usepackage{algorithmic}
\usepackage{graphicx}
\usepackage{textcomp}
\usepackage{xcolor}
\usepackage{multirow}
\def\BibTeX{{\rm B\kern-.05em{\sc i\kern-.025em b}\kern-.08em
    T\kern-.1667em\lower.7ex\hbox{E}\kern-.125emX}}
\begin{document}

\title{An End-to-End Network for Upright Adjustment of Panoramic Images}

\author{\IEEEauthorblockN{Heyu Chen\IEEEauthorrefmark{1}, Jianfeng Li\IEEEauthorrefmark{1} and Shigang Li \IEEEauthorrefmark{2}}
\IEEEauthorblockA{\IEEEauthorrefmark{1}College of Electronic and Information Engineering, Southwest University, Chongqing 400715, China}
\IEEEauthorblockA{\IEEEauthorrefmark{2}Graduate School of Information Sciences, Hiroshima City University, Hiroshima 731-3194, Japan}
Email: \{cy10123@email.swu.edu.cn, popqlee@swu.edu.cn, shigangli@hiroshima-cu.ac.jp\}
}

\maketitle

\begin{abstract}
Nowadays, panoramic images can be easily obtained by panoramic cameras. However, when the panoramic camera orientation is tilted, a non-upright panoramic image will be captured. Existing upright adjustment models focus on how to estimate more accurate camera orientation, and attribute image reconstruction to offline or post-processing tasks. To this end, we propose an online end-to-end network for upright adjustment. Our network is designed to reconstruct the image while finding the angle. Our network consists of three modules: orientation estimation, LUT online generation, and upright reconstruction. Direction estimation estimates the tilt angle of the panoramic image. Then, a converter block with upsampling function is designed to generate angle to LUT. This module can output corresponding online LUT for different input angles. Finally, a lightweight generative adversarial network (GAN) aims to generate upright images from shallow features. The experimental results show that in terms of angles, we have improved the accuracy of small angle errors. In terms of image reconstruction, In image reconstruction, we have achieved the first real-time online upright reconstruction of panoramic images using deep learning networks.
\end{abstract}


\section{Introduction}
\begin{figure}[htbp]
\centerline{\includegraphics{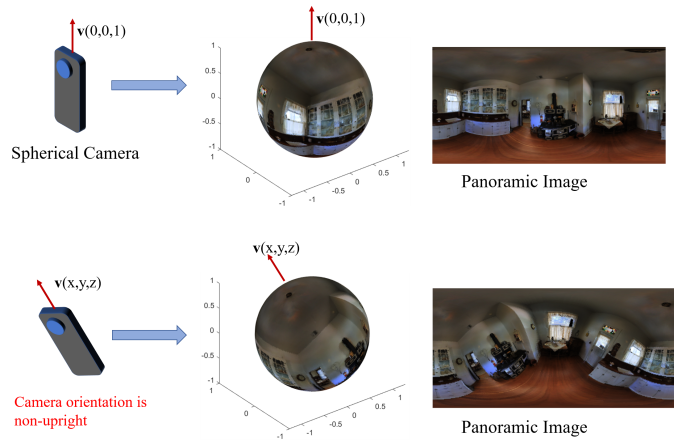}}
\caption{Use vector v to represent the camera direction. When the camera direction is upright, $\boldsymbol{v}=(0,0,1)$ (top left), and an upright spherical image is captured. When $\boldsymbol{v}=(x,y,z)\neq(0,0,1)$, camera orientation is nonupright, which captures nonupright spherical image, projected onto panoramic image with distortion (bottom right).}
\label{fig1}
\end{figure}

To realize immersive virtual/augmented reality, we often need to capture the surrounding scenes. The most convenient way to capture the surrounding scenes may be to use a spherical camera with 360 degrees field of view, as shown in Fig\ref{fig1}. The spherical images captured by such a spherical camera are also widely used in virtual tourism \cite{b1,b2} depth estimation \cite{b3,b4,b5,b6}, layout reconstruction \cite{b7}, cultural heritage \cite{b8}, recording crime scenes \cite{b9}, etc. To use a spherical image, we need to know its orientation; for example, when we extend a spherical image onto an equirectangular image, the appearance of the equirectangular image will change due to a different orientation (see Fig\ref{fig1}). Usually, similar to the research mentioned above, we assume that the spherical/equirectangular images are horizontal, which means that the horizon appears as a horizontal line in the images. For a spherical/equirectangular image captured with any orientation, we need to rectify it beforehand so that it has a canonical orientation. The processing is called the upright adjustment of panoramic images. 

In a more theoretical way, as Fig\ref{fig1} shows, we used a vector to represent the camera orientation, denoted by $\boldsymbol{v}$. When the camera orientation $\boldsymbol{v}$ is opposite to the direction of gravity, i.e.,$\boldsymbol{v}=(0,0,1)$, we can capture an upright spherical image. When the camera orientation is tilted, i.e.,$\boldsymbol{v}=(x,y,z)\neq(0,0,1)$, it will lead to a nonupright orientation of the captured image. That is, the canonical orientation is represented as $\boldsymbol{v}=(0,0,1)$, and the upright adjustment of panoramic images is to rotate an inclined image so that its orientation vector equals to $\boldsymbol{v}=(0,0,1)$. Since the canonical orientation can be obtained via a rotation with two rotational components, pitch and roll angles, the processing of upright adjustment of panoramic images involves two subtasks: rotational angle (pitch and roll) estimation and image remapping via rotation operation.

Recent studies on upright adjustment, either traditional feature-based methods or deep-learning-based methods, have until now focused on estimating the orientation of a panoramic image. Image reconstruction has not been much discussed and is usually processed offline using image remapping. An ideal situation would be to have a network that does upright adjustments directly without relying on post-processing.

To achieve this goal, in this paper, we propose an efficient end-to-end learning network to carry out upright adjustment of panoramic images, in which two subtasks, rotational angle (pitch and roll) estimation and image remapping via rotation operation, are completed by a single neural network. Our network is composed of three modules: orientation estimation, LUT online generation, and upright reconstruction. Our main contributions are as follows: 
\begin{itemize}
\item In contrast with the existing methods that output angles only and generate upright images offline, we are the first to propose a real-time online end-to-end solution to the best of our knowledge.
\item Our orientation estimation module has made some progress in the accuracy of small angle errors.
\item Upright reconstruction based on a lightweight cGAN is proposed to compensate for the error resulting from interpolation remapping to improve the fidelity of upright images.
\item As shown in the experiments, our proposed method can carry out upright adjustment of panoramic images online with about 11 fps and 429.6 MB storage. This work opens up the possibility of real-time upright adjustments on mobile devices.
\end{itemize}

The remain of this paper is organized as follows. In Section \ref{AA}, related works are introduced. In Section \ref{BB}, our proposed method is explained in details. After presenting the experimental results in Section \ref{CC}, conclusions are given in Section \ref{DD}.

\section{Related Works}\label{AA}
In this section, we introduce the related works, focusing on inclination estimation and Large-scale dataset generation.

\begin{figure*}[htbp]
\centerline{\includegraphics[width=0.95\textwidth]{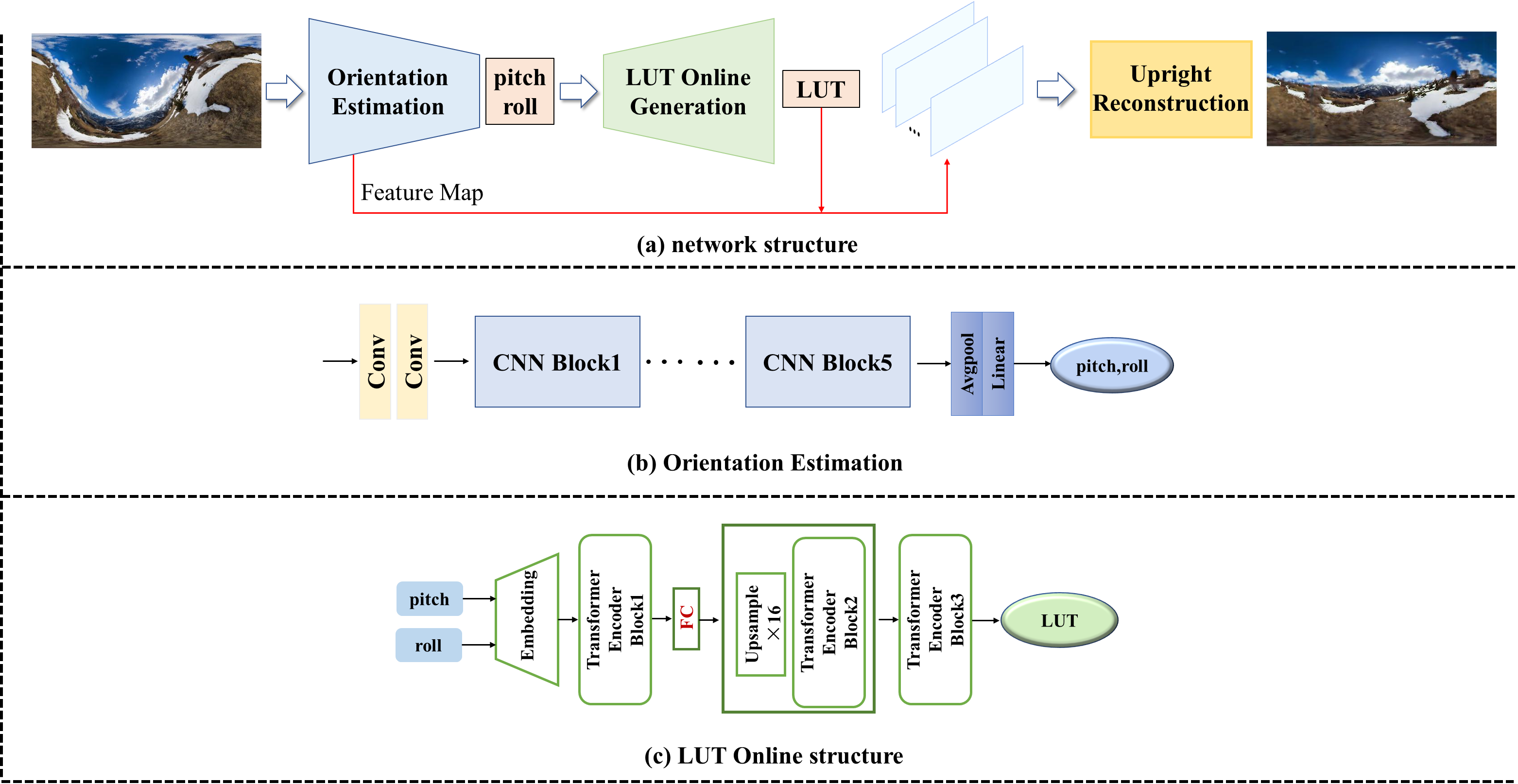}}
\caption{(a) Network structure overview diagram of our proposed method. (b) Orientation estimation model overview diagram. (c) Detail diagram of the LUT generation module. The details of transformer encoder block and upright reconstruction are further illustrated in Figures 3 and 4, respectively.}
\label{fig2}
\end{figure*}

\subsection{Inclination Estimation of Panoramic Images}\label{AA1}
The related works on inclination estimation of panoramic images can be divided into traditional feature-based methods and deep-learning-based methods.

In traditional feature-based methods base on the observation that vertical lines should appear vertical and the horizon should appear horizontal in upright adjusted panoramic images. Martins et al.\cite{b10} proposed a probabilistic sequential orientation estimation method based on a Manhattan world assumption likelihood model. However, there are many scenarios that do not satisfy the Manhattan world assumption. Demonceaux et al.\cite{b11,b12} estimated the horizon line by maximizing a criterion for sky/ground photometric separation and computing the upright orientation by finding the horizon line. Because this method relies on the horizon line, it is only for cases where there is a clear distinction between the sky and the ground. Bazin et al. \cite{b13,b14} proposed a line-based model and vanishing points model. Jung et al. \cite{b15} proposed an automatic method for upright adjustment of ${360}^{\circ}$ spherical panorama images without any prior information, such as depths and gyro sensor data. They take the Atlanta world assumption and use the horizontal and vertical lines in the scene to formulate a cost function for upright adjustment. The last two methods rely on the ability to distinguish between vertical structures and vanishing points. However, the apparent weak point of this kind of methods is that it is difficult to applied to natural landscape.

Recently, deep-learning-based methods are reported to carry out the inclination estimation of images. Fischer et al. \cite{b16} proposed using a convolutional network to estimate image orientation. Olmschenk et al. \cite{b17} proposed using convolutional neural networks (CNNs) to automatically determine the pitch and roll of a camera using a single, scene agnostic, 2D image. However, these two methods are not applied to panoramic images of upright adjustment. Joshi and Guerzhoy \cite{b18} apply convolutional neural networks (CNNs) to the problem of image orientation detection in the context of determining the correct orientation (from 0, 90, 180, and 270 degrees). Shima et al. \cite{b19} proposed an orientation detection method for face images that relies on image category classification by deep learning. Rotated images are classified into four classes, namely, ${0}^{\circ}$, ${90}^{\circ}$ clockwise, ${90}^{\circ}$ counterclockwise, or ${180}^{\circ}$. However, these two methods only estimate specific angles and only for perspective images. Jeon et al. \cite{b20} proposed an upright adjustment framework based on a CNN. Instead of directly predicting the 3D rotation of the camera on a given panorama image, their method estimates the rotation by analyzing the projected 2D rotations of multiple images sampled from the panorama. Although their study was a task of panorama estimation, it did not directly use the network to estimate the panorama angle. Jung et al. \cite{b21} proposed a method that consists of two modules (a CNN and a graph convolutional network (GCN)).

As for the state-of-the-art methods for upright adjustment of panoramic images in recent years. Deep360up \cite{b22}, uses DenseNet with a fully connected layer to predict the tilt direction of a spherical image, that is, the camera orientation v. The model was trained using a single angle loss, and the input of a panoramic image outputs the orientation. In \cite{b23}, the Coarse2Fine approach was proposed to divide the prediction of panoramic image orientation into two stages: the first stage adjusted the image within ${10}^{\circ}$, and the second stage refined the prediction accuracy within ${1}^{\circ}$. The output of this model was still the predicted angle, and the final predicted angle was affected by the first stage. In \cite{b24}, the camera orientation was predicted through the segmentation network combined with the vanishing point image and then adjusts the nonupright spherical image through the spherical rotation module. Although the above methods have greatly contributed to improving the accuracy of orientation estimation, they still divide the upright adjustment task into two steps: rotation angle estimation and panoramic image remapping. Their research only focuses on improving the accuracy of the angle estimation. 

\subsection{Large-Scale Dataset Generation }\label{AA2}
In the upright adjustment of panoramic images, there is no large dataset of nonupright panoramic images for deep learning training. Therefore, as early as in previous research, Jung et al. \cite{b22} proposed methods to generate large-scale datasets. The method is divided into three steps: 1) project the input equirectangular image onto the unit sphere 2) rotate the spherical image by fixed angles, 3) back-project the rotated spherical image back into the 2D plane. And throughout the process, to speed up the image remapping process. The above three steps can be accomplished by using pre-calculated LUTs (Look Up Tables). Since the camera rotation has only two degrees of freedom affecting the tilt, two angle parameters (pitch and roll are used in this article, the same as existing methods) are used to describe the rotation, i.e., an LUT for the remapping of the upright adjustment panoramic images is generated from two rotational angles. All LUTs will be obtained after determining the angle range and angle interval. Next use a random LUT to rotate the upright image to get a non-upright image. 

In the previous learning-based models, this method was used to generate dataset, so the method of generating dataset in this paper is consistent with it.

\section{End-to-end Learning Network for Panoramic Image Upright Reconstruction}\label{BB}

\begin{figure*}[htbp]
\centerline{\includegraphics[width=0.95\textwidth]{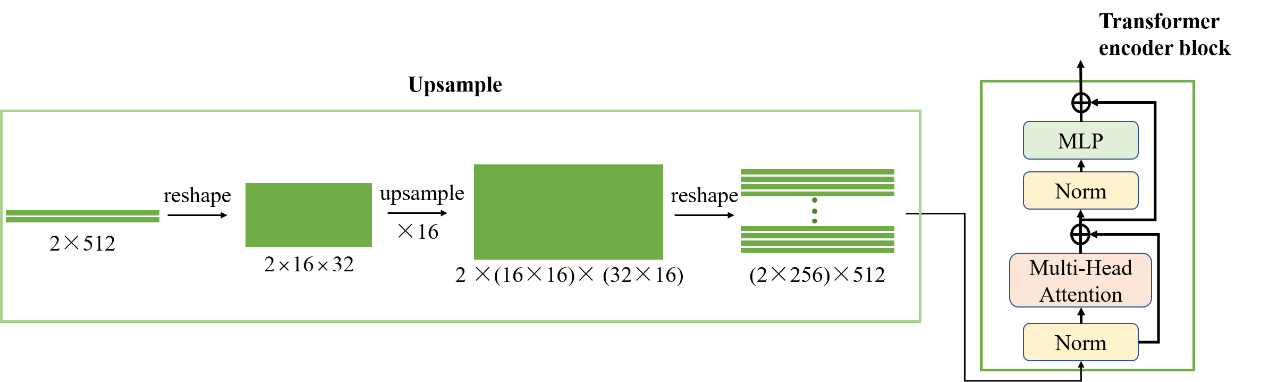}}
\caption{The details of upsample with transformer encoder block.}
\label{fig3}
\end{figure*}

Our proposed end-to-end learning network for panoramic image upright reconstruction consists of three modules: orientation estimation, LUT online generation, and upright reconstruction. The data flow direction of the entire network is shown in Fig\ref{fig2}(a). The nonupright image $I^{NUP}$ is input to the orientation estimation module to obtain pitch and roll. The whole process is represented as:
\begin{equation}
(roll,pitch)=A(I^{NUP})\label{eq1}
\end{equation}
where $A$ represents the orientation estimation module. Next, the estimated angles are input to the LUT generation module to obtain the corresponding 256x512 size LUT, i.e.:
\begin{equation}
LUT=G\_T(roll,pitch)\label{eq2}
\end{equation}
where $G\_T$ represents the LUT online generation module. 
Then, the nonupright shallow feature maps $F^{NUP}$ in the orientation estimation module are rotated with the generated LUT to obtain upright feature maps $F^{UP}$, i.e.:
\begin{equation}
F^{UP}=LUT(F^{NUP})\label{eq3}
\end{equation}

Finally, the upright feature maps are input to the upright reconstruction module to generate the final upright image $I^{UP}$ with high fidelity, i.e.:
\begin{equation}
I^{UP}=G\_I(F^{UP})\label{eq4}
\end{equation}
where $G\_I$ represents the image generation module. We will explain the details of the three modules in later sections.

\subsection{Orientation Estimation}\label{BB1}
Similar to other methods\cite{b22,b23,b24} for upright adjustment mentioned in Section\ref{AA1}, orientation estimation is an indispensable part of the upright adjustment task. Therefore, we designed an orientation estimation to provide input for the subsequent LUT generation module.

Section\ref{AA2} mentioned that the rotation for upright adjustment is determined by two degrees of freedom, so in this paper, the orientation estimation of the camera is transformed into pitch and roll. Previous work has demonstrated that convolutional neural networks can achieve orientation estimation, so it is reasonable to assume that features can be extracted by convolution, and the features naturally contain angle information. As the number of network layers increases, the network gradually obtains higher-level features, and the final angle is obtained by classifying the high-level features through the fully connected layer. Therefore, we design a pure convolution structure for estimating pitch and roll. The structure of the inclination estimation module is shown in Fig\ref{fig2}(b) and consists of five CNN blocks, an average pooling layer, and a fully connected layer. First, the image goes through two convolutions, which mainly represent the image with shallow features, such as some lines and contours, etc. These shallow features will be used as input in the subsequent upright reconstruction module.(Section\ref{BB3} for details) Then, the shallow features enter five CNN blocks, each of which consists of one max pooling and two convolutions. After each CNN block, the feature map size is reduced by half, and the number of channels is doubled. The input image size is 256x512, and the size of the feature map becomes 8x16 after 5 CNN blocks. The feature map goes through the average pooling layer and the fully connected layer to return two values between 0 and 1. These two values correspond to pitch and roll, respectively. Because the tilt angle generally does not exceed $[{-90}^{\circ},{90}^{\circ} ]$ for application, the pitch and roll of the predicted angles were calculated by $(pitch, roll) = (p, r) \times {180}^{\circ} - {90}^{\circ}$.

In this module, we use smooth L1 loss\cite{b25} to calculate the loss, considering that the difference between the true value and the predicted value is not greater than 1. To ensure that the gradient is not too small in the initial phase of training, the coefficient $\lambda$ is added. In summary, the angle loss is denoted as:
\begin{equation}
L_{Angle}=\lambda(0.5(p-\hat{p})^2-0.5(r-\hat{r})^2)\label{eq5}
\end{equation}
where $p$ represents the true value of pitch, $\hat{p}$ represents the predicted pitch of the model, $r$ represents the true value of roll, $\hat{r}$ represents the predicted roll of the model, and $\lambda=1000$.

\subsection{LUT Online Generation}\label{BB2}

\begin{figure*}[htbp]
\centerline{\includegraphics[width=0.95\textwidth]{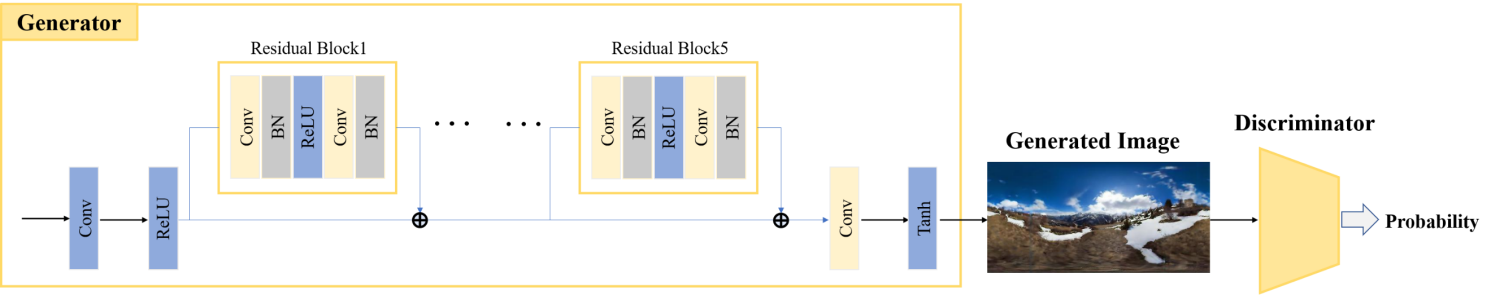}}
\caption{The details of upright reconstruction.}
\label{fig4}
\end{figure*}

We aim to propose an end-to-end solution to the upright alignment task, but in our practice we find that convolutions alone cannot directly generate upright panoramic images. That is, the nonlinear transformation from a non-upright image to an upright image on a two-dimensional plane cannot be easily learned by a convolutional network. Inspired by methods for generating large-scale datasets, location information that cannot be easily learned by the network can be provided by LUTs.

Although the LUT has good rotational efficiency, the LUT has the obvious disadvantage of requiring a large amount of memory. Because the size and number of LUT are determined by the image size and angle respectively. To balance the rotation efficiency and space occupancy, we propose a solution for generating LUTs. From the point of view of the traditional algorithm, the parameters of the LUT generation process are only roll and pitch, i.e., different LUTs are obtained by different roll and pitch. The size of the LUT is determined by the image/feature map that needs to be rotated. In this paper, we need to rotate the feature map of 256x512 size, so we need to complete the task of LUT generation of 256x512 size. Overall, a LUT generation of size 256x512 needs to be achieved. From the point of view of data volume, the estimated angle $a\in\mathbb{R}^2$ would be transformed to $lut\in\mathbb{R}^{2 \times 256 \times 512}$. Although a unique LUT can be obtained after determining the angle, it is very difficult for the network to learn the direct mapping relationship $G\_T:lut=G\_T(a)$. This difficulty stems from the fact that from $a\in\mathbb{R}^2$ to $lut\in\mathbb{R}^{2 \times 256 \times 512}$, the feature dimension span is too large, which will most likely lead to nonconvergence when training the network. Referring to the structure of CNN, which implemented downsampling followed by convolution to get the features, perhaps LUT online generation can be done in a similar reverse way to upsample the final 256x512 size LUT followed by feature encoder. Based on the above idea, our LUT online generation module can be divided into two steps: \begin{itemize}
\item Implementing $a\in\mathbb{R}^2$ to $lut\in\mathbb{R}^{2 \times 16 \times 32}$, a 16x32 LUT can provide a smooth transition.
\item Carry out upsampling with feature encoder: $\mathbb{R}^{2 \times 16 \times 32}$ to $\mathbb{R}^{2 \times 256 \times 512}$  . 
\end{itemize}

The LUT online generation network structure is shown in Fig\ref{fig2}(c). Next, we introduce the network details.

 \textbf{Embedding} As shown in Fig\ref{fig2}(c), input pitch and roll first into embedding. In fact, we use embedding to implement $a\in\mathbb{R}^2$ to $lut\in\mathbb{R}^{2 \times 16 \times 32}$. Embedding is used in natural language processing (NLP) to encode words and represent unstructured information as structured information. We embed pitch and roll values such as words in NLP, specifically embedding the two angles into two 1D vectors of 512 lengths, i.e.,
 \begin{equation}
a_i\in\mathbb{R}\xrightarrow{embedding}v_i\in\mathbb{R}^{512},i=1,2\label{eq6}
\end{equation}
 and the whole embedding process is learnable. Thus, during the training process, embedding is adjusted according to the loss to obtain a more suitable vector representation angle. Since the pitch and roll will obtain different LUTs if the values are exchanged, the angle position information $p_i$ is added after encoding the angle, where $i$ takes the value 1 or 2, corresponding to pitch and roll, respectively, and the final output of the network is obtained:
 \begin{equation}
E=\{z_1,z_2\}=\{v_1+p_1,v_2+p_2\}\label{eq7}
\end{equation}
$E$ is input into the transformer encoder block.

 \textbf{Transformer encoder block} The tremendous success of Transformer in natural language processing (NLP) has proven its superiority in processing sequences \cite{b27,b28}. Recent research has proven that the encoder of a transformer as the network backbone can have excellent performance in image generation tasks \cite{b29,b30}. In this paper, for the sequence $E$ obtained by embedding, our network backbone takes the encoder structure of transformer (Fig\ref{fig2}(c)) as well as \cite{b31,b32}. The transformer encoder block consists of two main sublayers, the multihead self-attention mechanism and the fully connection feed-forward network, with a residual connection \cite{b33} used around each of the two sublayers, followed by layer normalization \cite{b34}(the transformer encoder block details are shown on the right of Fig\ref{fig3}). In transformer encoder block1 of the module, the correlation between two vectors will be calculated by the multihead self-attention mechanism, and then the vectors will be reencoded by the residual connection. The correlation between the i and j vectors is denoted by $W_{i\_j}$, where $i$ and $j$ can only take the values 1 and 2 because our network has only two vectors used for input before upsampling (as shown in Fig\ref{fig2}(c), there are only two vectors obtained by embedding pitch and roll before upsampling). For the vectors $z_1$ and $z_2$ input into the multihead self-attention mechanism, the recoded $Z_1$ and $Z_2$ are denoted as:
 \begin{equation}
Z_1=z_1+W_{1\_1}(w_1z_1)+W_{1\_2}(w_2z_2)\label{eq8}
\end{equation}
\begin{equation}
Z_2=z_2+W_{2\_1}(w_1z_1)+W_{2\_2}(w_2z_2)\label{eq9}
\end{equation}
 where $w_1$ and $w_2$ are learnable parameters. $Z_1$ and $Z_2$ go through the feedforward network and the residual block to obtain the output of block 1. Although the two are represented as vectors related to each other, the output after these two vectors is still relatively independent. Since the LUT itself needs to be determined by two angles together and the angles are encoded into two vectors, that is, they need to be determined by two vectors together, we fuse the features of the two vectors through a layer of a fully connected layer (experiments show its necessity). At this time, the data volume of 2×512 is obtained after the first block. 
 
To generate a LUT of size 2×256×512, we upsample the data output from the first block. The details of the upsampling are shown in Figure 3. Reshape the sequence into 2×16×32 form before upsampling, and then perform ×16 upsampling. To maintain the consistency of the ransformer encoder, we reshape the representation back to the sequence. Two more transformer encoders are added for feature learning.

In the LUT online generation module, we use L1loss for training, considering that the range of values stored inside LUT is $(-1,1)$, so the final loss multiplication factor $\mu$ amplifies the loss value to help training. The loss is denoted as:
\begin{equation}
\begin{split}
L_{G\_T}&=\mu\frac{1}{2HW}\sum\left|LUT_t-LUT_{G\_T}\right|;\\
& LUT_t,LUT_{G\_T}\in\mathbb{R}^{2 \times H \times W}\label{eq10}
\end{split}
\end{equation}
where $LUT_t$,$LUT_{G\_T}$ denotes the real value of LUT and the generated value of LUT, respectively, and the coefficient $\mu=100$.

\subsection{Upright Reconstruction}\label{BB3}

\begin{table*}[htbp]
\caption{Performance of Different Methods in Sun360 Dataset}
\begin{center}
\begin{tabular}{|c|c|c|c|c|c|c|c|}
\hline

\multicolumn{2}{|c|}{}& \multicolumn{6}{c|}{Percentage of Predict Angle’s Deviation in Degrees}\\
\hline
Dataset& Method& $1^{\circ}$& $2^{\circ}$& $3^{\circ}$& $4^{\circ}$& $5^{\circ}$& ${12}^{\circ}$\\
\hline
\multirow{4}*{Sun360}& Ours& 29.9& \textbf{65.3}& \textbf{80.3}& 86.3& 89.2& 95.2\\
\cline{2-8}
&Segmentation\cite{b24}(2020)&19.7&53.6&75.5&\textbf{87.2}&\textbf{92.6}&\textbf{98.4}\\
\cline{2-8}
&Deep360up\cite{b22}(2019)&7.1&24.5&43.9&60.7&74.2&97.9\\
\cline{2-8}
&Coarse2Fine\cite{b23}(2019)&\textbf{30.9}&51.7&65.9&74.1&79.1&91.0\\

\hline
\end{tabular}
\label{tab1}
\end{center}
\end{table*}

To get the final upright image, we designed a third module (Fig\ref{fig3}). The input of the third module is the shallow feature map of the direction estimation module and the LUT of the LUT generation module. The shallow feature map is used to ease the learning difficulty of the generation module, which is consistent with the image size. For LUTs, in Section\ref{BB2}, we have obtained LUTs adapted to the size of feature maps.

The upright reconstruction module consists of a generator and a discriminator. The backbone network of the generator consists of five residual blocks \cite{b33}, which are used mainly to extract image features for the purpose of denoising. The structure of our discriminator comes from PatchGan \cite{b35}. The discriminator and the generator are alternately and iteratively trained. The purpose of the generator is to deceive the discriminator to generate a more realistic image. The probability judged by the discriminator is used as a loss to help the generator train. Overall, upright reconstruction is jointly trained with four losses. Specifically, the losses used by this module are as follows.

\textbf{Image Perceptual Loss}($L_{Perceptual}$) \cite{b26}. In this study, we use VGG19 to compute the features of the image, in which image perceptual loss is denoted as:
\begin{equation}
L_{Perceptual}=MSE((\varphi(\hat{X}),\varphi(X))\label{eq11}
\end{equation}
where $X$ denotes the upright panorama, $\hat{X}$ denotes the network generated image,  denotes the VGG19 feature extraction module, and $MSE()$ is used to calculate their mean-square error.

\textbf{SSIM Loss}($L_{ssim}$)\cite{b36}. The structural similarity index (SSIM) is proposed for measuring the structural similarity between images, based on independent comparisons in terms of luminance, contrast, and structures. The SSIM Loss is denoted as:
\begin{equation}
L_{ssim}=1-SSIM(\hat{X},X)\label{eq12}
\end{equation}

\textbf{Pixel Loss}($L_{pixel}$). The above two losses are used to match the training result parameters with human perception. In addition, we use L1 loss to constrain the image in pixel space: 
\begin{equation}
L_{pixel}=\left|\hat{X}-X\right|\label{eq13}
\end{equation}

\textbf{Discriminator Loss}($L_D$). As shown in Fig\ref{fig4}, the generated image is input to the discriminator, and the probability that the image is true is output. The discriminator loss is calculated as the binary cross-loss entropy of that probability with true. Experiments will show that the discriminator can solve the blurry in the upright reconstruction stage. The final joint loss is denoted as:
\begin{equation}
L=\alpha{L_{Perceptual}}+L_{ssim}+L_{pixel}+\beta{L_D}\label{eq14}
\end{equation}
where $\alpha=\beta=0.01$ and $L_D$ represents discriminator loss.

\section{Experiments}\label{CC}

\begin{figure*}[htbp]
\centerline{\includegraphics[width=0.95\textwidth]{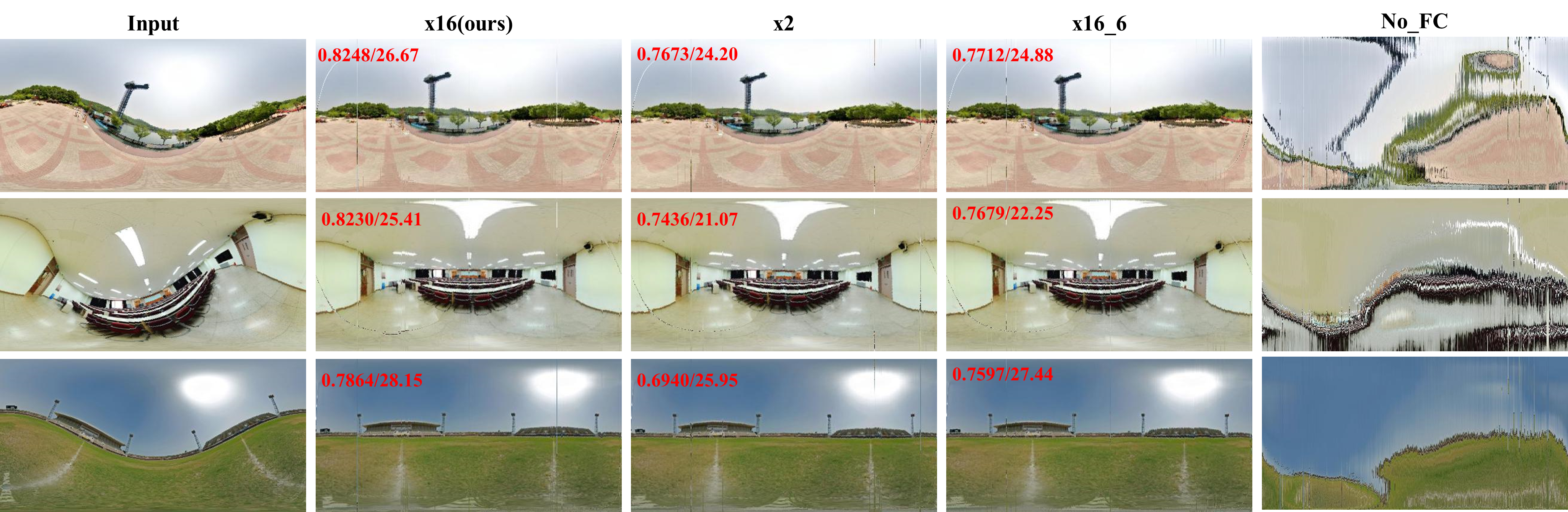}}
\caption{Ablation analysis of LUT Online Generation. x16: our proposed structure. x2: each layer is a x2 upsampling follow by a Transformer encoder. x16\_6: x16 upsampling follow by 6 transformer blocks. The red data in figures is ssim/psnr.}
\label{fig5}
\end{figure*}
\begin{figure*}[htbp]
\centerline{\includegraphics[width=0.95\textwidth]{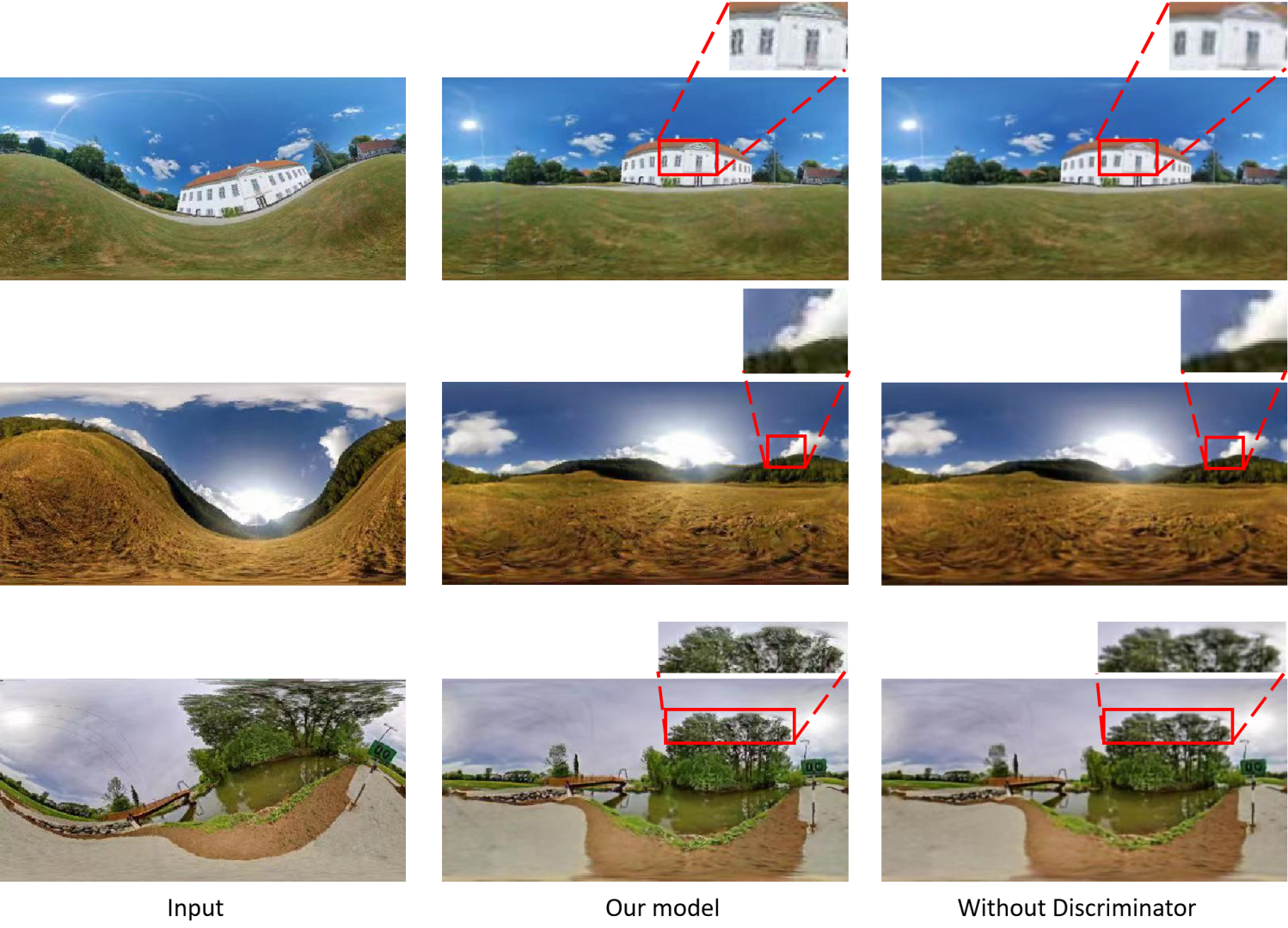}}
\caption{The first column is the input, the second column is the result of our model, and the third column is the result without the discriminator. }
\label{fig6}
\end{figure*}

\subsection{Dataset and Training Details}
Dataset In the deep-learning-based models of the previous research, the models all used the Sun360 dataset for training. Sun360 is a panoramic dataset and contains a variety of scenes, such as indoor, bridge building, and nature scenery scenes. Assume that the panoramic images in this dataset are all horizontal, i.e., $\boldsymbol{v}=(0,0,1)$ . The upright adjustment task requires nonupright panoramic images to train the network, so we needed to generate a dataset similar to the previous upright adjustment task. As mentioned in Section\ref{BB1}, the tilt range considered in this study was $[{-90}^{\circ},{90}^{\circ}]$, so to generate the dataset, we generated 181x181 LUT\_256x512 at 1° intervals in the $[{-90}^{\circ},{90}^{\circ}]$ range beforehand. It should be emphasized that LUT\_256x512 here was different from LUT mentioned in Section\ref{BB2}. LUT\_256x512 implements the function of rotating an upright image to a nonupright image for dataset preparation. Finally, in the Sun360 database (approximately 67,000 images), we use 70\% for training and 15\% and 15\% for testing and validation, respectively. We emphasize that since the Sun360 dataset contains multiple classifications, the Sun360 dataset is divided not simply by proportion but by ensuring the consistency of the classification images in the training, testing, and validation sets.

Because the LUT generation module requires 256x512 LUT truth values, similar to the above, we generate 181x181 LUTs at $1^{\circ}$ intervals in $[{-90}^{\circ},{90}^{\circ}]$, but the function of the LUT generation module is to rotate the nonupright image into an upright image, which is contrary to dataset preparation.

Training details All three modules of the network were trained on TITAN RTX 24G. The three modules of the network are trained separately in a certain order. First, the angle estimation module is trained using the Adam optimizer, a batch size of 8, a learning rate of $2 \times {10}^{-4}$, 40 epochs for Sun360 and one epoch that takes approximately 16 minutes. Load the angle estimation pretraining model in the subsequent training, freeze the parameters and no longer participate in the training. Training the LUT generator module using the stochastic gradient descent (SGD) optimizer, a batch size of 8, a learning rate of $3 \times {10}^{-2}$, 60 epochs for LUT and one epoch takes approximately 6 minutes. Similarly, the LUT generation module provides a pretrained model for the image generation module and does not update the parameters in the subsequent image generation training. After obtaining the pretrained models of the first two modules, the training of the upright reconstruction module starts, using the Adam optimizer, a batch size of 8, a generation learning rate of $2 \times {10}^{-4}$, a discriminator learning rate of ${10}^{-4}$, 4 epochs for Sun360 and one epoch takes approximately 45 minutes.

\subsection{Results of the orientation estimation module}
Since the input of the LUT online generation module is provided by the output of the orientation estimation module, to generate an accurate LUT, the accuracy of the angle cannot be ignored. Before presenting the model results, we briefly discuss how to evaluate the results of the orientation estimation module. When the camera orientation is kept upright, $\boldsymbol{v}=(0,0,1)$. When the camera orientation is tilted, $\boldsymbol{v}^\mathrm{T}=(x,y,z)^\mathrm{T}=R(p,r) \times (0,0,1)^\mathrm{T}$, where $p$ and $r$ are the truth values of roll and pitch. The network predicts the values of pitch and roll to be $\hat{p}$ and $\hat{r}$, respectively, and the camera orientation is $\boldsymbol{v_1}^\mathrm{T}=(x_1,y_1,z_1)^\mathrm{T}=R(\hat{p},\hat{r}) \times (0,0,1)^\mathrm{T}$. The angle between these two vectors ($\boldsymbol{v}=(x,y,z)$ and $\boldsymbol{v_1}=(x_1,y_1,z_1)$) is used to evaluate the error between the ground truth and the predicted value.

The results are shown in Table\ref{tab1}. To more intuitively observe the estimation results, the direction estimation results of other state-of-the-art methods are also listed in the table. While our method is only slightly lower than \cite{b24} in results with large angle errors, we've made progress on small angular errors and 95\% accuracy is sufficient for subsequent generation tasks from our proposed whole upright adjustment model. 

\subsection{Ablation analysis of LUT Online Generation}

In this section, we will discuss the structure of the LUT online generation module, including the necessity of a fully connected layer (FC), different upsampling strategies and numbers of encoder blocks. PSNR and SSIM are used for quantitative analysis.

To illustrate the importance of FC in the LUT online generation module, we remove the FC to retrain the network. As Fig\ref{fig5} (No\_FC) shows, the network does not converge in this situation, which proves the necessity of integrating the two angles in Section\ref{BB2}.

Considering the error caused by upsampling, upsampling layer by layer is tested, the network is designed as four layers, and each layer is a x2 upsampling followed by a transformer encoder. The result is shown in Fig\ref{fig5} (x2). The proposed structure in Section\ref{BB2} is shown in Fig\ref{fig5} (x16). According to PSNR and SSIM, the current x16 upsampling is the best.

To test the effect of encoder block numbers, we test x16 upsampling followed by 6 transformer blocks. Fig\ref{fig5} (x16\_6) shows that it did not improve the effect. In our opinion, the LUT generation task is not as complicated as common image generation tasks; more layers may decrease the training efficiency, and a simple network is sufficient for our task.

\subsection{Evaluating discriminator in upright reconstruction}
In this section, we will show the results of our model and the model after removing the discriminator. The experimental results are shown in Fig\ref{fig6}. In Fig\ref{fig6}, we find that the model without the discriminator has a certain degree of blur, such as the tree texture, cloud and house.

\subsection{Results for Upright Adjustment with Our Network}

\begin{figure*}[htbp]
\centerline{\includegraphics[width=0.95\textwidth]{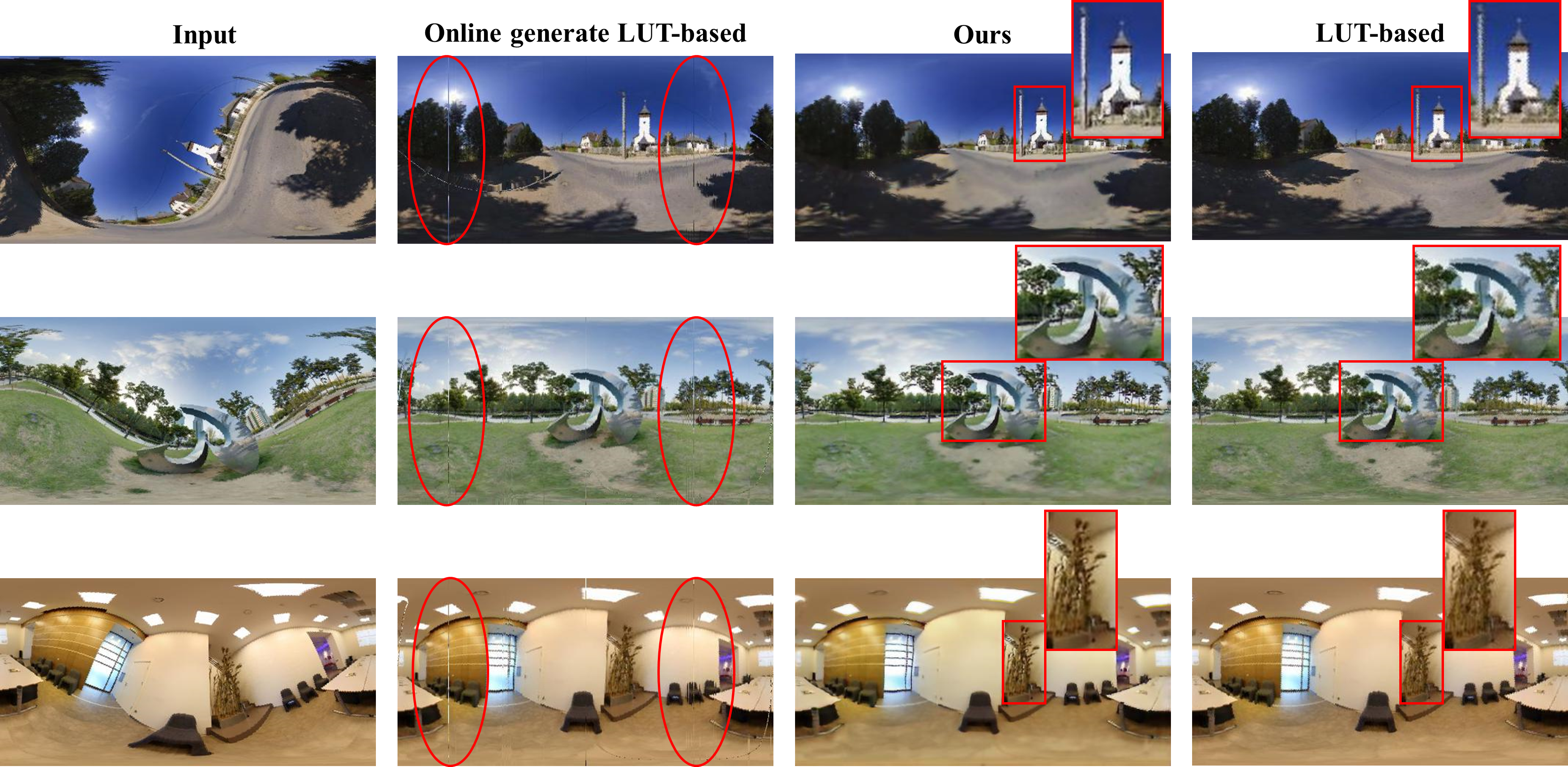}}
\caption{Presentation of the input, the image from LUT generation module (with a certain of noise), the image from upright construction module, and the image rotated by the ground truth LUT}
\label{fig7}

\end{figure*}
\begin{figure*}[htbp]
\centerline{\includegraphics[width=0.95\textwidth]{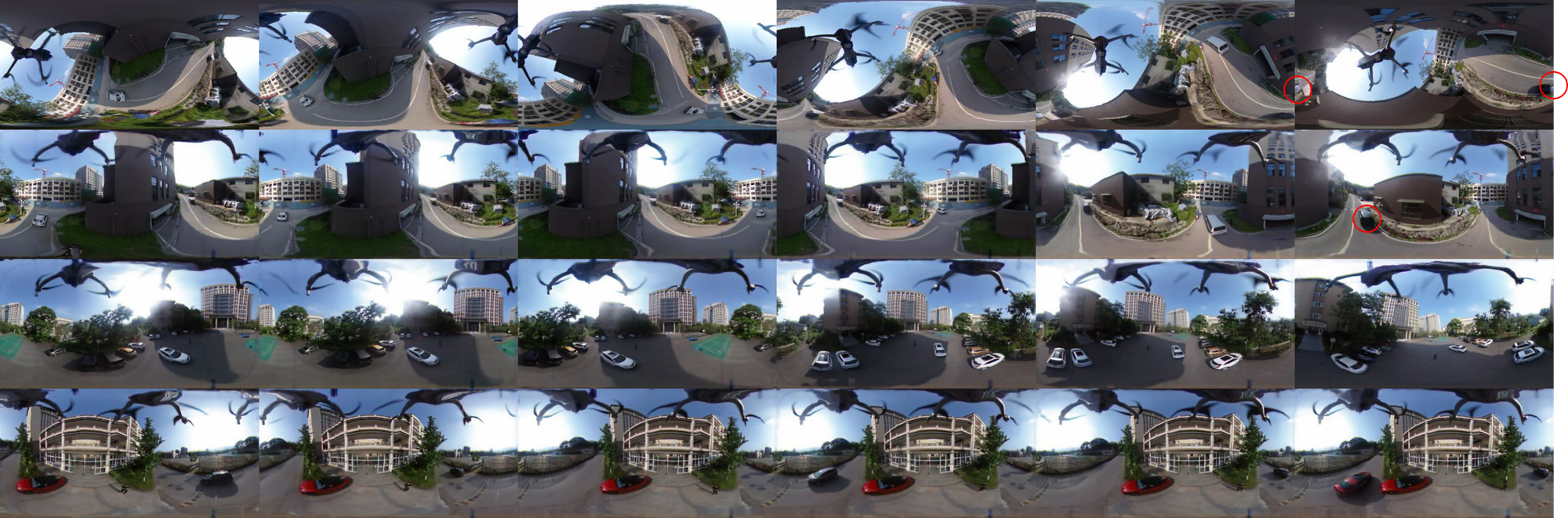}}
\caption{The first column is the input, the second column is the result of our model, and the third column is the result without the discriminator.}
\label{fig8}
\end{figure*}

In this section, we introduce our upright adjustment models in three aspects: time, space, and generated image quality. 
Time  We upright resize the entire test set of images on the server using our trained model, and the time measurement part excludes the image loading and preprocessing part. Measures the time from the beginning of the image entering the model to the end of the image output, that includes orientation estimation, LUT online generation and upright reconstruction. We measured an average spin time of 0.012 seconds. Obviously our model operation meets the requirements of real-time processing, which provides the possibility to realize the upright adjustment on the line.

Spatially Section\ref{BB2} mentioned that by using LUT to help the network generate upright images, according to the LUT rotation image principle, the LUT should contain all possibilities so that it can be corrected for any rotation angle of the scene. For the $[{-90}^{\circ},{90}^{\circ}]$ range considered in our model, we needed a total of 32761 (181x181) LUTs. and the size of the LUTs has to be kept in line with the image size, while in our model, we use 256x512 size images, Thus taking up 4.65 GB of space in our server's storage environment. While our final network (i.e., including orientation estimation, LUT online generation, upright reconstruction) only takes up 429.6 MB. Therefore it is very necessary to use the LUT generation module to optimize the storage occupied by the LUT, and 429.6MB reduces the storage requirements and can be suitable for more devices.

Image quality The upright images generated by our network are shown in Fig\ref{fig7}. The objects we compared are images rotated directly using the LUT. From the point of view of the current upright adjustment methods, LUT-based is the second step they did not mention: use direct rotation or LUT-adjust for nonupright images. Since the current work only discusses how to improve the accuracy of the estimation and the actual method used to adjust the image is a default action (direct rotation or use LUT), we focus on the comparison with the traditional LUT-based. In Fig\ref{fig7}, the image rotated with our generated LUT will have noise (the second column of Fig\ref{fig7}), which is why we designed the upright reconstruction module. That is, the upright reconstruction module not only completes the task of generating images from feature maps but also denoises them. The LUT-based has a jagged edge due to the interpolation of remapping (enlarged in Fig\ref{fig7}). The image we generated is relatively less jagged because it has been processed by the network and also has a certain elimination effect on the fixed noise caused by the LUT.

\subsection{Panorama Reconstruction}
To test our method in real situations, we took a sequence of real images with a drone carrying a commercial panoramic camera. Commercial panoramic cameras usually record the orientation by the gyroscope inside and then upright the panoramic images on computers offline. Supposing that the images directly acquired by the embedded system are at any orientation, without any adjustment. They should be considered the first row in Fig\ref{fig8}, and the operator should try to observe the scene from the first-person perspective in real time. Obviously, these pictures cannot be transmitted to the unmanned aerial vehicle (UAV) observer since they are not user-friendly. If these images are fed into our end-to-end online network, a real-time rotation can be achieved Finally, all the images are horizontal (the last three rows in Fig\ref{fig8}). The observer of this UAV would have no feeling that the drone has tiled due to some reasons, which brings better quality of continuity.

Furthermore, our online method accelerated the practicability of other panoramic image tasks. As we know, most panoramic image tasks require an upright image, and they cannot be used online if the adjustment step is offline. Now, their network could be connected directly followed by our network. For example, we can see that the last picture in the first line of Fig\ref{fig8} with a red circle, there is a black vehicle inside. The black vehicle should be horizontal in the real world but here vertical in images, since the camera is tilted. It would be difficult to identify, but it would be much easier after our method. After all, if online adjustment of panoramic images can be realized, it is quite significant for other real-time tasks.
\section{Conclusions}\label{DD}
We proposed an online upright adjustment method for panoramic images. The greatest feature is the proposed online LUT generation module, which helps the convolutional network to generate upright images to achieve online upright adjustment. In addition, to achieve the upright adjustment task, we also design an upright reconstruction module to reduce the error of interpolation remapping. Furthermore, among existing models, we are the first to propose an online upright adjustment model, and our network achieves competitive image quality while achieving upright adjustment in real time, opening up the possibility that this task can be used in embedded systems.

\vspace{12pt}
\end{document}